\documentclass[letterpaper]{article} % DO NOT CHANGE THIS
\usepackage{aaai2026}  % DO NOT CHANGE THIS
\usepackage{times}  % DO NOT CHANGE THIS
\usepackage{helvet}  % DO NOT CHANGE THIS
\usepackage{courier}  % DO NOT CHANGE THIS
\usepackage[hyphens]{url}  % DO NOT CHANGE THIS
\usepackage{graphicx} % DO NOT CHANGE THIS
\usepackage{natbib}  % DO NOT CHANGE THIS AND DO NOT ADD ANY OPTIONS TO IT
\usepackage{caption} % DO NOT CHANGE THIS AND DO NOT ADD ANY OPTIONS TO IT
\usepackage{algorithm}
\usepackage{algpseudocode}

\usepackage{newfloat}
\DeclareCaptionStyle{ruled}{labelfont=normalfont,labelsep=colon,strut=off} % DO NOT CHANGE THIS
\usepackage{tikz}
\usetikzlibrary{arrows}
\usetikzlibrary{shapes.geometric, intersections}
\tikzset{
	treenode/.style = {align=center, inner sep=0pt, text centered, font=\sffamily}, arn_n/.style = {treenode, circle, white, font=\sffamily\bfseries, draw=black, text width=1.5em},% arbre rouge noir, noeud noir
	arn_r/.style = {treenode, rectangle, draw=black, fill=black, minimum width=0.5em, minimum height=0.5em},% arbre rouge noir, noeud rouge
	arn_x/.style = {treenode, rectangle, draw=black, minimum width=0.5em, minimum height=0.5em},% arbre rouge noir, nil
	arn_e/.style = {treenode, circle, black, font=\sffamily\bfseries, draw=black, text width=1.5em},% arbre rouge noir, noeud noir
}

\usepackage{adjustbox}
\usepackage{amsmath,amsthm,amssymb}
\usepackage{siunitx}
\usepackage{import}

\usepackage{pgf}
\usepackage{pgfplots}
\usepgfplotslibrary{units}
\pgfplotsset{compat = newest}

\everymath=\expandafter{\the\everymath\displaystyle}

\theoremstyle{definition}

\title{Monte Carlo Permutation Search}
\author{
Tristan Cazenave
}
\affiliations{
LAMSADE, Université Paris Dauphine - PSL, Paris, France}

\begin{document}

\maketitle

\begin{abstract}
We propose Monte Carlo Permutation Search (MCPS), a general-purpose Monte Carlo Tree Search (MCTS) algorithm that improves upon the GRAVE algorithm. MCPS is relevant when deep reinforcement learning is not an option or when the computing power available before play is not substantial, such as in General Game Playing. The principle of MCPS is to include in the exploration term of a node the statistics on all the playouts that contain all the moves on the path from the root to the node. We test MCPS on a variety of games: Hex, Go, AtariGo, NoGo and a Wargame. MCPS almost always outperforms GRAVE. We also provide a mathematical derivation of the formulas used for weighting the three sources of statistics. These formulas are an improvement on the GRAVE formula since they no longer use the bias hyperparameter of GRAVE.
\end{abstract}

\begin{figure*}[t]
\centering
\begin{tikzpicture}[
  mv/.style={draw, rounded corners=1.5pt, minimum width=0.6cm,
             minimum height=0.6cm, font=\small, inner sep=1pt},
  path/.style ={mv, fill=blue!20,   draw=blue!60,   text=blue!50!black},
  act/.style  ={mv, fill=orange!40, draw=orange!85, text=orange!55!black},
  other/.style={mv, fill=gray!10,   draw=gray!50,   text=gray!40!black},
  ttl/.style={font=\small\bfseries, anchor=west},
  lbl/.style={font=\scriptsize\itshape, anchor=west, text=gray!50!black},
  newlbl/.style={font=\scriptsize\bfseries, anchor=west, text=red!70!black}
]

% Header (centered over the whole figure: midpoint of x range [-11.5, 4.5] is -3.5)
\node[font=\bfseries] at (-3.5, 3.7)
  {Evaluating action $a$ at node $s$, reached by path $(a_0, a_1, a_2)$};

% Row 1: Q
\node[ttl] at (-11.5, 2.8) {$Q(s,a)$};
\node[lbl] at (-9.7,  2.8) {playout begins with the path, then $a$};
\node[path]  at (-1.5, 2.8) {$a_0$};
\node[path]  at (-0.8, 2.8) {$a_1$};
\node[path]  at (-0.1, 2.8) {$a_2$};
\node[act]   at ( 0.6, 2.8) {$a$};
\node[other] at ( 1.3, 2.8) {$x$};
\node[other] at ( 2.0, 2.8) {$y$};
\node[other] at ( 2.7, 2.8) {$z$};
\node[font=\small] at ( 3.3, 2.8) {$\cdots$};

% Row 2: Q tilde
\node[ttl] at (-11.5, 1.9) {$\tilde Q(s,a)$};
\node[lbl] at (-9.7,  1.9) {begins with the path; $a$ appears later (AMAF, used by GRAVE)};
\node[path]  at (-1.5, 1.9) {$a_0$};
\node[path]  at (-0.8, 1.9) {$a_1$};
\node[path]  at (-0.1, 1.9) {$a_2$};
\node[other] at ( 0.6, 1.9) {$x$};
\node[other] at ( 1.3, 1.9) {$y$};
\node[act]   at ( 2.0, 1.9) {$a$};
\node[other] at ( 2.7, 1.9) {$z$};
\node[font=\small] at ( 3.3, 1.9) {$\cdots$};

% Row 3: Q hat
\node[ttl] at (-11.5, 1.0) {$\hat Q(s,a)$};
\node[newlbl] at (-9.7, 1.0)
  {contains $\{a_0, a_1, a_2, a\}$ in any order, anywhere \textmd{\itshape (new in MCPS)}};
\node[other] at (-1.5, 1.0) {$x$};
\node[path]  at (-0.8, 1.0) {$a_2$};
\node[act]   at (-0.1, 1.0) {$a$};
\node[path]  at ( 0.6, 1.0) {$a_0$};
\node[other] at ( 1.3, 1.0) {$y$};
\node[path]  at ( 2.0, 1.0) {$a_1$};
\node[other] at ( 2.7, 1.0) {$z$};
\node[font=\small] at ( 3.3, 1.0) {$\cdots$};

% Separator
\draw[gray!40] (-11.7, 0.4) -- (4.5, 0.4);

% Combined formula
\node[ttl] at (-11.5, -0.2) {MCPS selection rule:};
\node[font=\small] at (-3.5, -0.4)
  {$\displaystyle \mathrm{val}(s,a) \;=\;
    \frac{n\, Q(s,a) \;+\; \tilde n\,\tilde Q(s_r,a) \;+\; \hat n\,\hat Q(s_r,a)}
         {n \;+\; \tilde n \;+\; \hat n}$};

% Legend
\node[path]  at (-7.0, -1.4) {};
\node[anchor=west, font=\scriptsize] at (-6.7, -1.4) {path move};
\node[act]   at (-4.0, -1.4) {};
\node[anchor=west, font=\scriptsize] at (-3.7, -1.4) {candidate action $a$};
\node[other] at (-0.5, -1.4) {};
\node[anchor=west, font=\scriptsize] at (-0.2, -1.4) {other move};

\end{tikzpicture}
\caption{The MCPS contribution. To evaluate an action $a$ at node $s$ reached by path $(a_0, a_1, a_2)$, GRAVE uses two estimators: $Q$ from playouts that start with the path and $a$, and $\tilde Q$ from playouts that start with the path with $a$ appearing later. MCPS adds a third, $\hat Q$, which pools all playouts containing the path moves and $a$ as a multiset --- in any order and at any positions. The three estimators are then combined with weights proportional to their sample counts, the variance-minimizing convex combination under independence, removing GRAVE's bias hyperparameter.}
\label{fig:mcps-contribution}
\end{figure*}

\section{Introduction}

Monte Carlo Tree Search (MCTS) \citep{Kocsis2006,Coulom2006} has been successfully applied to many games and problems \citep{BrownePWLCRTPSC2012,swiechowski2023monte}. It originates from the computer game of Go \citep{Bouzy01} with a method based on simulated annealing \citep{Bruegmann1993MC}. The principle underlying MCTS is to learn which move to play using statistics from random games. In the early times of MCTS, random games were played with a uniform policy. Computer Go programs soon used non-uniform playout policies, learning the policy with optimization algorithms \citep{Coulom2007}. Playout policies were replaced with neural network evaluations for computer Go with the AlphaGo program \citep{Silver2016MasteringTG}, and then for other games such as Chess and Shogi with the AlphaZero program \citep{silver2018general}. There have been numerous applications of MCTS following its notable success in Computer Go. Applications to problems other than games include interplanetary trajectory planning \citep{hennes2015interplanetary}, real-world disaster response \citep{wu2015agile}, agile legged locomotion \citep{clary2018monte}, multi-robot active perception \citep{best2019dec},    security games \citep{karwowski2020double}, chemical retrosynthesis \citep{segler2018planning,genheden2020aizynthfinder,roucairol2024comparing,blackshaw2025enhancing}, fluid-structure topology optimization \citep{gaymann2019deep}, task scheduling \citep{hu2019spear}, vehicle routing \citep{mandziuk2017mcts,sentuc2023learning}, and general problem solving \citep{sabar2015population,Cazenave15}, among others.

Algorithms that use MCTS with playouts instead of evaluation by a trained neural network are useful for domains such as General Game Playing \citep{Pitrat68,Genesereth2005} when the game is not known in advance, or for other applications, such as modeling biological systems where the biological system varies with each run \citep{michelucci2024improving}. We aim to develop a general MCTS algorithm that performs well across many problems without parameter tuning, without training a model as in AlphaZero, and without problem-specific modifications. GRAVE \citep{Cazenave15} is such an algorithm, and we propose to compare it to others in our experiments.

In a recent study \citep{soemers2024towards} related to a Kaggle competition, 268,386 plays were performed among 61 different agents across 1,494 distinct games. The goal of the Kaggle competition was to design the best machine learning model to predict the strength of general MCTS algorithms using different optimizations. A result of the Kaggle competition is that the GRAVE algorithm is the best feature for predicting the strength of general MCTS agents.

We propose to improve on GRAVE using three sources of statistics instead of two, as in GRAVE. The first two sources are the same as in GRAVE: the statistics on the playouts that start with a move and the AMAF statistics on the playouts below a node that contain a move. The third source of statistics is on the playouts that contain a sequence of moves in any order. The algorithm is named Monte Carlo Permutation Search (MCPS) as it uses statistics on playouts for which a permutation of the moves of the playout starts with the same sequence of moves as the sequence of moves that reaches the state associated with the node in the search tree. For some games, such as Hex, any permutation of the moves of a playout results in the same state as if we had directly played the permuted sequence. For other games, such as AtariGo, if we replay a permutation, it can lead to a state different from the state at the end of the original playout with an opposite score. Nevertheless, we have found that even for AtariGo, the use of statistics on permutations improves the winrate. An explanation is that capture moves happen only once in a game as the winning move. In order to take this into account, we added the capture feature in the AtariGo codes for moves, thus recovering the permutation validity. Figure \ref{fig:mcps-contribution} summarizes the contribution of the paper.

This paper is organized as follows. The second section describes previous work. The third section presents the combination of statistics used in MCPS and the MCPS algorithm. The fourth section details the experimental results. The last section concludes.

\section{Previous Work}

The first program to produce statistics on random games in the game of Go was the Gobble program \citep{Bruegmann1993MC}. It also defined the AMAF statistics as a heuristic to use when there are not enough playouts.

MCTS \citep{Coulom2006} was then designed by memorizing the visited states in a tree and using them to direct the search with the UCB exploration term, leading to the UCT algorithm \citep{Kocsis2006}. It was a revolution in computer game playing.

The same year, Virtual Global Search \citep{cazenave2006virtual} was proposed. It used statistics on playouts containing a sequence of moves played in any order. These statistics are similar to the permutation statistics we propose to use in combination with usual statistics and AMAF statistics.

An important improvement to UCT was the RAVE algorithm \citep{gelly2011monte}. It combined AMAF statistics with the usual statistics. It was a significant improvement over UCT for computer Go.

The use of transposition tables in UCT was also studied with the Upper Confidence Bound for rooted Directed acyclic graphs (UCD) algorithm \citep{SaffidineCazenaveMehat2011KBS}.

The GRAVE algorithm \citep{Cazenave15} is a simple modification of RAVE that makes it much better than RAVE for many games. It uses the AMAF statistics of an ancestor node instead of the AMAF statistics of the node, as in RAVE. It is an appropriate algorithm for General Game Playing. It was recently adapted to the continuous case for modeling biological systems \citep{michelucci2024improving}.

MCTS is often used in combination with Deep Learning since the success of the AlphaGo system \citep{Silver2016MasteringTG} and subsequent versions such as AlphaZero \citep{silver2018general}. The exploration term used by AlphaGo is PUCT which makes use of the prior on the moves given by the policy head. AlphaZero has been applied to numerous problems, including algorithm discovery for matrix multiplication \citep{fawzi2022discovering} and quantum circuit optimization \citep{ruiz2025quantum}, for example.

\section{Combining Statistics}

In this section, we start by defining the notations for the various statistics used in the paper. We then detail the mathematical derivation of the formulas for the weights applied to the different statistics. We continue with the description of the MCPS algorithm.

\subsection{Notations}

Let a state $s$ be defined as the sequence of $d$ actions from the root that reaches $s$: 

$$s = a_0, a_1, a_2, ..., a_d$$

Let a playout $p$ be defined as the sequence of its actions until the terminal state:

$$p = p_0, p_1, p_2, ..., p_t$$

%Let $\euscr{P}$ be the set of all playouts and 
Let $P(s)$ be the set of playouts that start from $s$. Let $avg(P(s))$ be the function that returns the average of the rewards of the playouts in the set of playouts $P(s)$.

We can now define:

$$Q(s,a) = avg (\{p \in P(s)~|~p_0 = a \})$$

$$n(s,a) =  |\{p \in P(s)~|~p_0 = a\}|$$

$$n(s) =  |\{p \in P(s)\}|$$

and $\tilde Q(s,a)$, the All Moves As First (AMAF) statistics at node s:

$$\tilde Q(s,a) = avg (\{p \in P(s)~|~a \in p \})$$

$$\tilde n(s,a) =  |\{p \in P(s)~|~a \in p\}|$$

%Let $\hat s (r)$ be the deepest state from the root to $s$ that has more than $r$ playouts.

%$$\hat Q(s,a,r) = \tilde Q(\hat s (r),a)$$

%$$\hat N(s,a,r) =  \tilde N(\hat s (r),a)$$

Let $\hat Q(s,a)$ be the permutation statistics at node $s$ that take into account all the playouts containing the moves of $s$ in any order, with $root$ as the root of the search tree:

$$\hat Q(s,a) = avg (\{p \in P(root)~|~a \in p, \forall a_i \in s : a_i \in p\})$$

$$\hat n(s,a) =  |\{p \in P(root)~|~a \in p, \forall a_i \in s : a_i \in p\}|$$

\subsection{Mathematical Derivation of the Formulas for the Weights of the Different Statistics}

%Let $Q_*(s,a)$ be the weighted sum of $Q (s,a), \tilde Q(s,a), \text{and } \hat Q (s,a)$:

Let define:

\[
Q_*(s,a) = \beta Q (s,a) + \tilde \beta \tilde Q(s_r,a) + \hat \beta \hat Q (s_r,a)
\]

\text{with } $$\beta + \tilde \beta + \hat \beta = 1$$

Assume the variances are:
\[
\mathrm{Var}(Q) = \frac{\sigma^2}{n}, \quad
\mathrm{Var}(\tilde{Q}) = \frac{\sigma^2}{\tilde{n}}, \quad
\mathrm{Var}(\hat{Q}) = \frac{\sigma^2}{\hat{n}}.
\]

The variance of the combined estimator is:
\[
\sigma^2 (\frac{\beta^2}{n} + \frac{\tilde{\beta}^2}{\tilde{n}} + \frac{\hat{\beta}^2}{\hat{n}}).
\]

Let
\[
N = n + \tilde{n} + \hat{n}.
\]
Let the simplified variance be defined as follows:
\[
V = \frac{\beta^2}{n} + \frac{\tilde{\beta}^2}{\tilde{n}} + \frac{\hat{\beta}^2}{\hat{n}}.
\]

We consider the candidate weights:
\[
\beta = \frac{n}{N}, \quad
\tilde{\beta} = \frac{\tilde{n}}{N}, \quad
\hat{\beta} = \frac{\hat{n}}{N}.
\]

With these weights, the simplified variance becomes:
\[
V^* =
\frac{n^2}{N^2 n}
+ \frac{\tilde{n}^2}{N^2 \tilde{n}}
+ \frac{\hat{n}^2}{N^2 \hat{n}}
= \frac{1}{N}.
\]

Now consider any other weights:
\[
\beta' = \beta + x, \quad
\tilde{\beta}' = \tilde{\beta} + y, \quad
\hat{\beta}' = \hat{\beta} + z,
\]
with the constraint:
\[
x + y + z = 0.
\]

The new simplified variance is:
\[
V' =
\frac{(\beta + x)^2}{n}
+ \frac{(\tilde{\beta} + y)^2}{\tilde{n}}
+ \frac{(\hat{\beta} + z)^2}{\hat{n}}.
\]

Expanding:
\[
V' =
V^*
+ \frac{x^2}{n}
+ \frac{y^2}{\tilde{n}}
+ \frac{z^2}{\hat{n}}
+ 2x \frac{\beta}{n}
+ 2y \frac{\tilde{\beta}}{\tilde{n}}
+ 2z \frac{\hat{\beta}}{\hat{n}}.
\]

Since
\[
\frac{\beta}{n} = \frac{1}{N}, \quad
\frac{\tilde{\beta}}{\tilde{n}} = \frac{1}{N}, \quad
\frac{\hat{\beta}}{\hat{n}} = \frac{1}{N},
\]
the linear terms cancel:
\[
2x\frac{1}{N} + 2y\frac{1}{N} + 2z\frac{1}{N}
= \frac{2}{N}(x+y+z) = 0.
\]

Thus,
\[
V' = V^* + \frac{x^2}{n} + \frac{y^2}{\tilde{n}} + \frac{z^2}{\hat{n}} \ge V^*.
\]

Therefore, the choice
\[
\beta = \frac{n}{n+\tilde{n}+\hat{n}}, \quad
\tilde{\beta} = \frac{\tilde{n}}{n+\tilde{n}+\hat{n}}, \quad
\hat{\beta} = \frac{\hat{n}}{n+\tilde{n}+\hat{n}}
\]
minimizes the variance when assuming the statistics are independent. The statistics are not independent in practice, since the playouts used to calculate $Q$ are also used to calculate $\tilde Q$ and $\hat Q$. We will nevertheless use these formulas.

\subsection{The MCPS Algorithm}

MCPS is a Monte Carlo Tree Search algorithm that combines two complementary bandit priors at every selection step. The first is the \emph{GRAVE} prior~\cite{Cazenave15}, which supplies AMAF-style statistics from the nearest sufficiently-visited ancestor. The second is the \emph{permutation prior} $\hat{Q}$, which counts, within a sliding window of recent playouts, how many contain the same move codes as the current tree path and what outcomes they obtained.

\paragraph{Selection.}
At each tree node $s$, move $a$ carries a direct visit count $n$ and a win sum $w$, giving $Q (s,a) = w / n$ (defaulting to $1$ when $n = 0$). Two additional estimates are drawn from ancestor nodes:

\begin{itemize}
\item $\tilde{n}$ and $\tilde{Q} (s_r,a)$: the AMAF count and winrate for move $a$ at the reference node $s_r$.
\item $\hat{n}$ and $\hat{Q} (s_r,a)$: the permutation count and winrate for move $a$ at the reference node $s_r$.
\end{itemize}

The selected child maximizes
\begin{equation}\label{eq:mcps-grave}
    \mathrm{val}(a)
    \;=\;
    \frac{n Q (s,a) + \tilde{n}\tilde{Q} (s_r,a)
        + \hat{n}\hat{Q} (s_r,a)}
         {n + \tilde{n} + \hat{n}}
  \end{equation}
negated for the opponent's turn. A move with $\tilde{n}=0$ is given priority over all others.

\paragraph{GRAVE prior.}
Each node $s$ accumulates AMAF counters $(\tilde{n}_c,\tilde{w}_c)$ for every code $c$ that appears anywhere in the playouts that pass through $s$. After each playout, the first occurrence of each code in the played sequence updates the AMAF counters of every node on the current tree path.  The GRAVE reference $s_r$ for a node is the nearest ancestor (inclusive) on the path whose visit count exceeds a threshold $\rho$; nodes borrow that ancestor's richer statistics until their own count exceeds $\rho$.

\paragraph{Bitsets.}
Each move code $c$ maintains a compact bitset $B_c$ over a sliding window of $W$ recent playouts (default $W=10{,}000$): bit $p \bmod W$ is set if playout $p$ contains code $c$, and the corresponding outcome $r_p$ is stored in a parallel array.  When a playout is replaced by a newer one, its bit and its outcome are updated.

%At the \emph{root}, the permutation statistics are unconditional: every playout in the window is relevant, so
%  \[
%    \hat{n}_c = |B_c|,
%    \qquad
%    \hat{Q}_c = \frac{\displaystyle\sum_{p\in B_c} r_p}{|B_c|}.
%  \]

\paragraph{Computation of permutation statistics.}
When a non-root node $s$ first accumulates $n = \rho$ visits, its permutation statistics are computed and frozen for all future visits. The permutation statistics are computed in a single pass over 64-bit words using \textsc{popcount} and bit-scan instructions. The node stores $(\hat{n}_c,\,\hat{Q}_c)$ for all codes. Node $s$ then becomes the new permutation reference $s_r$ for its entire subtree.

\paragraph{Propagation to descendant nodes.}
Once a node $s$ has frozen its permutation statistics, the reference $s_r = s$ is passed down to every recursive call in its subtree. Any descendant that has not yet reached $\rho$ visits reads $\hat{n}_c$ and $\hat{Q}_c$ directly from $s_r$'s cache, and uses these in Equation~\eqref{eq:mcps-grave}. When a descendant $d$ itself reaches $\rho$ visits, it freezes its own statistics and becomes the new permutation reference for $d$'s subtree.

\paragraph{Move codes.}
Each legal move is mapped to an integer code that captures game-relevant features: the moving player, the destination cell, and, for the Wargame, whether the move triggers combat.  Statistics are keyed by code rather than exact moves, enabling knowledge transfer between distinct board positions.

\begin{algorithm}[h!]
\begin{algorithmic}[1]
\State $s \leftarrow root$,  $s_r \leftarrow root$
\While{$s$ is in the tree and is not terminal}
    \If{$n(s) = \rho$}
        \For{each $a \in \text{moves}$}
            \State Calculate and freeze $\hat Q (s,a)$
        \EndFor
    \EndIf
    \If{$n(s) \geq \rho$}
        \State $s_r \leftarrow s$
    \EndIf
    \For{each $a \in \text{moves}$}
        \State $Q_*(s,a) \leftarrow \beta Q (s,a) + \tilde{\beta} \tilde Q(s_r,a) + \hat{\beta} \hat Q (s_r,a) $
    \EndFor
    \State Select $a_* \leftarrow \operatorname{argmax}\{Q_*(s, a) \mid a \in \text{moves}\}$
    \State $s \leftarrow s \cup \{a_*\}$
\EndWhile
\State Add $s$ to the tree as the new leaf
\While{$s$ is not a terminal state}
    \State Sample $a$ from the available moves of $s$
    \State $s \leftarrow s \cup \{a\}$
\EndWhile
\State score $\leftarrow \text{evaluate}(s)$
\State Update $Q, N, \tilde Q,$ and $\tilde N$ for nodes on the path from the root to the new leaf 
\State $P(\mathit{root}) \leftarrow P(\mathit{root}) \cup s$;\quad update cyclic bitsets $\{B_c\}$
\end{algorithmic}
\caption{\label{MCPS}The MCPS algorithm. $B_c$ is the bitset of recent playouts (cyclic window) that contain move code~$c$.
}
\end{algorithm}

\begin{algorithm}
  \caption{\label{Freeze}Freeze $\hat{Q}$ statistics at node $s$, called when $N(s) = \rho$. The key step is line 2–4: ANDing the bitsets for every code on the path $a_0,\ldots,a_d$ gives exactly the playouts that contain the codes to node $s$ via that path. Intersecting $V$ with $B_{c_a}$ (line 6) then restricts to those that also played action $a$ from $s$, yielding the path-conditioned sample for $\hat{Q}(s,a)$.}
  \begin{algorithmic}[1]
  \Require Path codes $a_0, a_1, \ldots, a_d$ from root to $s$;
           bitset $B_c$ (playout indices in the cyclic window that contain code $c$);
           playout results $r_i$
  \Ensure  $\hat{n}(s,a)$ and $\hat{Q}(s,a)$ stored for every child action $a$

  \State $V \leftarrow B_{a_0}$
  \For{$k \leftarrow 1$ \textbf{to} $d$}
      \State $V \leftarrow V \wedge B_{a_k}$
      \Comment{$V[i]=1$ iff playout $i$ contains $a_0,\ldots,a_k$}
  \EndFor
  \For{each possible code $c_a$}
      \State $W \leftarrow V \wedge B_{c_a}$
      \State $\hat{n}(s,a) \leftarrow \lvert W \rvert$
      \If{$\hat{n}(s,a) > 0$}
          \State $\hat{Q}(s,a) \leftarrow \displaystyle\sum_{i \,:\, W[i]=1} r_i\;/\;\hat{n}(s,a)$
      \Else
          \State $\hat{Q}(s,a) \leftarrow 1$ \Comment{optimistic default}
      \EndIf
  \EndFor
  \end{algorithmic}
\end{algorithm}

The MCPS algorithm is Algorithm \ref{MCPS}. The algorithm to compute the permutation statistics is Algorithm \ref{Freeze}.

\section{Experimental Results}

In our implementation of GRAVE and MCPS, we use a transposition table to store information about the states in the search tree \citep{SaffidineCazenaveMehat2011KBS}. This means that the permutations of moves in the tree are already naturally handled. The permutations considered with MCPS deal with the entire playout's permutations, not just the permutations of the path to the node that have already been accounted for with the transposition table.

For two-player games and for each parameter configuration, we run 800 games between MCPS and GRAVE: 400 games with MCPS playing first and 400 games with GRAVE playing first. With 800 games, the 95\% confidence interval half-width is roughly ±3.5 percentage points. MCPS and GRAVE always use the same wall clock time during their search. 

All of the experiments are conducted by running 100 processes in parallel. Each process is assigned a different random seed. All the tables are presented by board size (rows) and per-move time budget (columns).

For all the experiments, we used a fixed $\rho = 30$ and a cyclic playout bitset capped at 10,000 playouts.

\subsection{Hex}

\begin{table}[ht]
\centering
\begin{tabular}{llrrrr}
\hline
Game & Size & 0.25s & 0.5s & 1s & 2s \\
\hline
Hex & $5\times 5$ & 43.12 & 45.88 & 46.62 & 49.25 \\
Hex & $7\times 7$ & 53.00 & 49.00 & 55.62 & 60.25 \\
Hex & $9\times 9$ & 64.00 & 70.50 & 67.50 & 63.62 \\
Hex & $11\times 11$ & 68.12 & 68.25 & 72.12 & 76.25 \\
Hex & $13\times 13$ & 64.25 & 73.25 & 72.00 & 76.12 \\
Hex & $15\times 15$ & 61.75 & 66.62 & 71.88 & 71.88 \\
\hline
\end{tabular}
\caption{Winrate of MCPS against GRAVE on Hex. Hex is permutation-invariant. MCPS outperforms GRAVE from $7\times 7$ upward, with the largest gains on boards of $11\times 11$ and above; on $5\times 5$ the bitset overhead is not amortised and MCPS falls slightly below GRAVE.}
\label{tab:grave_mcps_hex}
\end{table}

Hex is the cleanest setting for MCPS: any permutation of the moves of a playout reaches the same terminal state, so the permutation statistics are unconditionally valid without any code-feature workaround.

Table~\ref{tab:grave_mcps_hex} reports the winrates of MCPS against GRAVE. The results are size-dependent: MCPS is below $50\%$ on $5\times 5$ at all four time budgets, climbs rapidly to the mid-60s on $9\times 9$, and plateaus in the low-to-mid 70s from $11\times 11$ to $15\times 15$. Hex is the only game in our study where MCPS underperforms GRAVE in any cell, and the underperformance is confined to the smallest board.

The appendix tables show a corresponding speed pattern. On small boards, MCPS is actually slower than GRAVE: at $5\times 5$, MCPS runs around $85\%$ of GRAVE's playouts per second (Tables~\ref{tab:grave_pps_hex} and~\ref{tab:mcps_pps_hex}). The cost of maintaining the playout bitsets is a fixed overhead per playout, and on $5\times 5$ where playouts are very short ($\sim 4$ moves) and few moves are visited, the overhead dominates. As the board grows, the picture reverses: on $15\times 15$ at $1$s, MCPS runs at $1.36\times$ GRAVE's speed because MCPS builds deeper trees that absorb a larger fraction of the game and leave shorter playouts because the win is reached in fewer moves (Tables~\ref{tab:grave_len_hex} and~\ref{tab:mcps_len_hex}). On $15\times 15$ at $2$s, GRAVE playouts average $58.5$ moves while MCPS playouts average $40.2$ --- a $31\%$ reduction in simulation work per iteration.

The depth gain is monotonic in absolute terms across budgets and grows with the board size in relative terms (Tables~\ref{tab:grave_dep_hex} and~\ref{tab:mcps_dep_hex}). At $2$s, MCPS reaches leaves $4\%$ deeper than GRAVE on $5\times 5$ but $33\%$ deeper on $15\times 15$. This translates directly into wins: Hex has no early-termination mechanism analogous to capture, so additional tree depth always carries the search further into the relevant decision space.

\subsection{Go}

\begin{table}[ht]
\centering
\begin{tabular}{llrrrr}
\hline
Game & Size & 0.25s & 0.5s & 1s & 2s \\
\hline
Go & $5\times 5$ & 55.12 & 55.50 & 49.88 & 50.00 \\
Go & $7\times 7$ & 57.75 & 55.88 & 54.12 & 52.50 \\
Go & $9\times 9$ & 59.50 & 58.12 & 59.25 & 60.62 \\
Go & $11\times 11$ & 61.88 & 63.25 & 62.62 & 62.12 \\
Go & $13\times 13$ & 58.38 & 60.12 & 63.00 & 63.75 \\
Go & $15\times 15$ & 59.25 & 62.25 & 60.38 & 63.00 \\
\hline
\end{tabular}
\caption{Winrate of MCPS against GRAVE on Go. MCPS outperforms GRAVE in every cell from $7\times 7$ upward and across all budgets from $9\times 9$ upward, with the winrate gap on the larger boards widening as the time budget grows.}
\label{tab:grave_mcps_go}
\end{table}

Go is played under Chinese rules with simple ko and a komi of 6.5 points for White.

Table~\ref{tab:grave_mcps_go} reports the winrates of MCPS against GRAVE. The size dependence and the time-budget dependence both matter here, and they pull in opposite directions on the small boards. On $5\times 5$ and $7\times 7$, MCPS leads at the $0.2$s budget ($55$--$58\%$) but the gap shrinks as time grows, and $5\times 5$ at $1$s and $2$s sits right at break-even. From $9\times 9$ upward, the trend reverses: the MCPS advantage is stable or grows with time, reaching $63$--$64\%$ on $13\times 13$ and $15\times 15$ at the $2$s budget. Go is the only game in the study where MCPS's advantage clearly widens with thinking time on the larger boards.

The appendix data shows where this comes from. The two algorithms run at very similar playout rates on Go --- MCPS is between $0.96$ and $1.17$ times GRAVE's playouts per second (Tables~\ref{tab:grave_pps_go} and~\ref{tab:mcps_pps_go}) --- and MCPS playouts are only modestly shorter than GRAVE's, typically by $5$--$15\%$ (Tables~\ref{tab:grave_len_go} and~\ref{tab:mcps_len_go}). The decisive difference is in how each algorithm spends additional thinking time on building its tree (Tables~\ref{tab:grave_dep_go} and~\ref{tab:mcps_dep_go}). MCPS's new-leaf depth grows steadily with the time budget at every board size: going from $0.2$s to $2$s, MCPS deepens its tree by a factor of $1.24$ on $7\times 7$, $1.27$ on $13\times 13$, and $1.61$ on $15\times 15$. GRAVE behaves very differently: the average leaf depth does not increase as much with more time; it even decreases in some cases. The branching factor in Go is high, and with more playouts, GRAVE appears to spend the extra budget exploring shallow alternatives rather than committing depth to a promising line.

The consequence is that the absolute depth advantage of MCPS over GRAVE widens dramatically with thinking time. On $11\times 11$, MCPS reaches leaves $1.8$ deeper than GRAVE at $0.2$s but $7.1$ deeper at $2$s. On $13\times 13$, the two algorithms are essentially tied in depth at $0.2$s ($+0.1$) but MCPS is $6.6$ ahead at $2$s. This matches the winrate trend on the larger boards: extra time disproportionately benefits MCPS, which is why the winrate on $13\times 13$ climbs from $58.38$ at $0.2$s to $63.75$ at $2$s. On the small boards, the story is different: both algorithms eventually find good moves, so MCPS's depth advantage — though still present — stops translating into wins as time grows and the game approaches saturation.

\subsection{AtariGo}

\begin{table}[ht]
\centering
\begin{tabular}{llrrrr}
\hline
Game & Size & 0.25s & 0.5s & 1s & 2s \\
\hline
AtariGo & $5\times 5$ & 64.50 & 69.00 & 68.38 & 60.25 \\
AtariGo & $7\times 7$ & 60.75 & 66.50 & 68.88 & 67.50 \\
AtariGo & $9\times 9$ & 54.12 & 55.75 & 62.88 & 68.38 \\
AtariGo & $11\times 11$ & 54.12 & 54.00 & 55.75 & 58.50 \\
AtariGo & $13\times 13$ & 52.75 & 54.50 & 52.25 & 54.00 \\
AtariGo & $15\times 15$ & 55.00 & 52.38 & 54.38 & 54.00 \\
\hline
\end{tabular}
\caption{Winrate of MCPS against GRAVE on AtariGo. MCPS outperforms GRAVE in every cell, with the largest gains on small boards.}
\label{tab:grave_mcps_AtariGo}
\end{table}

AtariGo is a capture-the-first-stone variant of Go: the first player to capture an opponent's stone wins. As noted in the introduction, replaying a permutation of an AtariGo playout does not, in general, reach the same terminal state, since a capture move terminates the game. We accommodate this by including a capture flag in the move code, so that capturing and non-capturing moves on the same cell are treated as distinct events by the permutation statistics.

Table~\ref{tab:grave_mcps_AtariGo} reports the winrates of MCPS against GRAVE across board sizes and time budgets. MCPS outperforms GRAVE in every cell, with the largest gains on small boards: $5\times 5$ and $7\times 7$ reach winrates in the high 60s, while $13\times 13$ and $15\times 15$ settle in the low-to-mid 50s. This is the opposite trend from Hex, Go, and NoGo, where the advantage of MCPS grows with board size.

The appendix tables clarify the picture. MCPS and GRAVE run at essentially the same number of playouts per second across all sizes (Tables~\ref{tab:grave_pps_AtariGo} and~\ref{tab:mcps_pps_AtariGo}), with MCPS slightly faster in most configurations --- the bitset-based permutation statistics add no measurable overhead. MCPS playouts are consistently shorter than GRAVE's (Tables~\ref{tab:grave_len_AtariGo} and~\ref{tab:mcps_len_AtariGo}), and MCPS reaches deeper leaves in the tree across all board sizes (Tables~\ref{tab:grave_dep_AtariGo} and~\ref{tab:mcps_dep_AtariGo}), with the relative gain in tree depth roughly uniform across sizes (around $10$--$25\%$ deeper at the $2$s budget).

The winrate trend is therefore not explained by where MCPS searches more effectively --- it searches more effectively everywhere --- but by how close that search reaches to terminal states. On $5\times 5$ AtariGo, playouts average around four moves, and MCPS reaches leaves at depth ten, covering most of the decision-relevant horizon; the improved statistics translate directly into wins. On $15\times 15$, playouts average around 23 moves while leaves are added at depth four, leaving most of the game beyond the search frontier and diluting the benefit of better in-tree selection.

\subsection{NoGo}

\begin{table}[ht]
\centering
\begin{tabular}{llrrrr}
\hline
Game & Size & 0.25s & 0.5s & 1s & 2s \\
\hline
NoGo & $5\times 5$ & 61.75 & 62.50 & 63.88 & 63.62 \\
NoGo & $7\times 7$ & 60.12 & 61.00 & 65.12 & 62.75 \\
NoGo & $9\times 9$ & 59.62 & 66.25 & 65.12 & 67.00 \\
NoGo & $11\times 11$ & 60.75 & 63.50 & 62.75 & 69.50 \\
NoGo & $13\times 13$ & 59.12 & 61.25 & 63.25 & 65.25 \\
NoGo & $15\times 15$ & 57.38 & 62.00 & 61.00 & 64.00 \\
\hline
\end{tabular}
\caption{Winrate of MCPS against GRAVE on NoGo. MCPS outperforms GRAVE in every cell, with winrates ranging from $57.4\%$ to $69.5\%$ and a per-size average between $61$ and $65\%$.}
\label{tab:grave_mcps_NoGo}
\end{table}

NoGo is a combinatorial game played on an $n\times n$ board under Go rules, with the exception that all captures are forbidden: a move is illegal if it would remove an opponent group's last liberty (capture forbidden) or leave the placed group without any liberty (suicide forbidden). The player with no legal move loses; there is no komi and no passing.

Table~\ref{tab:grave_mcps_NoGo} reports the winrates of MCPS against GRAVE. NoGo is the most uniformly favorable game in the study for MCPS: every cell is between $57.4\%$ and $69.5\%$, with the per-size average between $61\%$ and $65\%$. Unlike Hex (where MCPS loses on $5\times 5$), AtariGo (where the gap shrinks on large boards), or the Wargame (where the gap is modest on the small boards), NoGo shows no regime in which MCPS underperforms or saturates. The winrate also drifts gently upward with the time budget on most sizes, but the effect is much weaker than what we observed on Go.

The appendix tables show that the underlying mechanism is a modest improvement in each of the three quantities we measure rather than a strong effect in any single one. MCPS runs $1.7$--$6.5\%$ more playouts per second than GRAVE across all sizes (Tables~\ref{tab:grave_pps_NoGo} and~\ref{tab:mcps_pps_NoGo}); MCPS playouts are $3$--$10\%$ shorter than GRAVE's (Tables~\ref{tab:grave_len_NoGo} and~\ref{tab:mcps_len_NoGo}); and MCPS reaches leaves $14$--$36\%$ deeper than GRAVE (Tables~\ref{tab:grave_dep_NoGo} and~\ref{tab:mcps_dep_NoGo}). The depth gain grows with the time budget at every board size.

\subsection{Wargame}

\begin{table}[ht]
\centering
\begin{tabular}{llrrrr}
\hline
Game & Size & 0.25s & 0.5s & 1s & 2s \\
\hline
Wargame & $5\times 5$ & 55.88 & 51.12 & 53.00 & 57.75 \\
Wargame & $7\times 7$ & 55.50 & 52.12 & 49.38 & 51.75 \\
Wargame & $9\times 9$ & 59.38 & 60.50 & 58.88 & 58.38 \\
Wargame & $11\times 11$ & 59.38 & 59.25 & 60.12 & 61.62 \\
Wargame & $13\times 13$ & 57.88 & 57.88 & 57.88 & 61.88 \\
\hline
\end{tabular}
\caption{Winrate of MCPS against GRAVE on the Wargame. MCPS outperforms GRAVE from $9\times 9$ upward.}
\label{tab:grave_mcps_wargame}
\end{table}

We designed a two-player asymmetric strategy game played on an $N{\times}N$ grid with hex adjacency. We experiment with $N \in \{5,7,9,11,13\}$. Each player controls $N{-}1$ units, all starting with two health points.

\paragraph{Setup phase.}
Players alternate placing units one at a time. Red (the attacker) places in columns $0$ and $1$; Blue (the defender) places in columns $N{-}2$ and $N{-}1$. The phase ends once all $2(N{-}1)$ units are on the board.

\paragraph{Movement phase.}
Players alternate turns. On each turn, the active player selects one alive unit and moves it to an adjacent empty cell (hex distance~1). Red's winning cell---Blue's base located at $(r,c) = (\lfloor N/2 \rfloor,\, N{-}1)$---is the sole exception: Red may enter it even when Blue occupies it, displacing the occupant.

\paragraph{Combat.}
Immediately after a unit moves, it automatically attacks the weakest adjacent enemy (lowest HP; ties broken by the enemy's hex distance to  Blue's base). The attacked unit loses one HP; a unit reduced to zero HP is removed from the board.

\paragraph{Codes for moves.}
Each move in the Wargame is identified by the unit that moves and its destination cell, but for the purpose of AMAF and permutation statistics, we abstract away unit identity and encode only the three attributes that are strategically relevant: the active player, the destination cell, and whether the move triggers an attack. This scheme groups together all moves that land on the same cell with the same strategic consequence, allowing statistics gathered from one unit's trajectory to inform the evaluation of the same destination reached by a different unit.

\paragraph{Terminal conditions.}
Red wins if any of its units enter Blue's base, or if all Blue units are eliminated. Blue wins if all Red units are eliminated, or if the game reaches $4N^2$ half-moves without Red winning. The move limit prevents draws from cycling.

\paragraph{Characteristics.}
The game is highly asymmetric: Red must break through a defensive line across an open hex field, while Blue must intercept and attrit the  attackers. With $N=9$ there are $8$ units per side, a $9{\times}9$  board, and a move limit of $324$ half-moves. The branching factor  during the movement phase reaches up to $6(N{-}1)$ per turn.

\paragraph{Playout Policy.}
The playout policy for the Wargame is a biased random simulation that reflects the strategic objectives of each side. During the placement phase both players assign unit positions uniformly at random among their legal starting columns. During the movement phase a weighted distribution over legal moves is used for each player. Red's weight for moving unit $u$ from cell $f$ to cell $t$ is $1 + 2\max(0, d(f,g) - d(t,g))$, where $g$ is the target cell and $d$ denotes Manhattan distance; moving directly onto $g$ receives an exceptional weight of $100$. Blue's weight for unit $u$ is $1 + 2\max(0, d(f,r_u) - d(t,r_u))$, where $r_u$ is the position of the nearest surviving Red unit to $u$'s current cell $f$. In both cases a move that brings the unit no closer to its objective receives weight $1$, ensuring that all legal moves retain a positive probability of being selected. A move is then sampled from the resulting discrete distribution using a thread-local Mersenne Twister. This heuristic reflects the core dynamic of the game: Red rushes toward the target while Blue attempts to intercept the most advanced Red unit.

\paragraph{Experiments.}
Table~\ref{tab:grave_mcps_wargame} reports the winrates of MCPS against GRAVE in the Wargame across board sizes and time budgets. The results are split into two regimes. On the smallest boards ($5\times 5$ and $7\times 7$), the gap between MCPS and GRAVE is modest and depends on the time budget, with winrates oscillating in the low-to-mid 50s; the $7\times 7$ board at $1$s sits at $49.38\%$, within the confidence interval of break-even. From $9\times 9$ upward, MCPS consistently reaches the high 50s to low 60s across all four time budgets.

The appendix tables show that the Wargame is the most favorable game in our study for MCPS in raw computational terms. MCPS runs $15$ to $33\%$ more playouts per second than GRAVE across all board sizes (Tables~\ref{tab:grave_pps_wargame} and~\ref{tab:mcps_pps_wargame}). This speedup comes from substantially shorter playouts: MCPS playouts run at $78$--$91\%$ of GRAVE's length, with the reduction growing on larger boards (Tables~\ref{tab:grave_len_wargame} and~\ref{tab:mcps_len_wargame}). The tree itself is also a little deeper, with MCPS reaching leaves $8$--$15\%$ deeper than GRAVE (Tables~\ref{tab:grave_dep_wargame} and~\ref{tab:mcps_dep_wargame}); the depth gain is fairly uniform across board sizes and budgets.

The most striking effect is on the total length of a simulated game (tree depth plus playout length). On $5\times 5$ the two algorithms simulate games of comparable length ($\sim 25$ moves), but on $13\times 13$ MCPS simulated games average around $60$ moves while GRAVE simulated games average around $72$ --- the games end roughly $17\%$ earlier under MCPS. Neither algorithm comes close to the $4N^2$ move limit at any size, so this is not a truncation effect: MCPS appears to converge on decisive engagements (a breakthrough by Red or an extermination by Blue) faster than GRAVE. This combination of more playouts per second, shorter playouts, and deeper trees explains why the winrate advantage stabilizes in the high 50s to low 60s from $9\times 9$ upward, even though the per-cell margins on the smallest boards remain modest.

\section{Conclusion}

We presented MCPS, a Monte Carlo Tree Search algorithm that extends GRAVE with a third source of statistics: the win rates of playouts containing the moves on the path from the root to the current node, in any order. We derived analytically the formula for the weighting of the three sources, obtained as the convex combination that minimizes the variance of the combined estimator under independence. This formula has no tunable bias hyperparameter, unlike the formula used by GRAVE.

Experiments on five games --- Hex, Go, AtariGo, NoGo, and an asymmetric Wargame --- show that MCPS outperforms GRAVE on every game when averaged across configurations, with per-game and per-budget patterns that vary in interpretable ways. NoGo is the most uniformly favorable case, with winrates between $57\%$ and $69\%$ across all board sizes and time budgets. On Hex, the advantage of MCPS grows with the board size. On AtariGo, the trend is reversed: MCPS's gains are largest on small boards, because there the search reaches close to terminal states, and shrink on large boards where most of the game remains beyond the search frontier. On Go, the advantage of MCPS widens markedly with the time budget on the larger boards: MCPS deepens its tree steadily as more time becomes available, while GRAVE's new-leaf depth on the same boards is non-monotonic and sometimes decreases. On the Wargame, MCPS runs $15$ to $33\%$ more playouts per second than GRAVE and converges on decisive engagements faster, with games ending around $17\%$ earlier under MCPS on $13\times 13$.

These patterns are consistent with a single underlying mechanism: MCPS produces a better-informed selection rule, which lets it build deeper trees and, depending on the game, either reduce simulation work, reach closer to terminal states, or convert extra time into deeper exploration of promising lines. The benefit appears when the resulting depth advantage falls within the relevant decision horizon of the game.

The idea of combining three sources of statistics in MCTS is not specific to MCPS and can be reused for other kinds of statistics. We hope that MCPS will be used for problems beyond those described in this paper, as it is a general MCTS algorithm.

%\section*{Acknowledgment}

\newpage

\bibliography{main}

\appendix

\section*{Appendix}

\subsection{AtariGo}

The following tables give the per-cell measurements underlying the AtariGo experiments. For each board size and time budget, we report the number of playouts completed per second (Tables~\ref{tab:grave_pps_AtariGo} and~\ref{tab:mcps_pps_AtariGo}), the average number of moves in a playout from the new leaf to a terminal state (Tables~\ref{tab:grave_len_AtariGo} and~\ref{tab:mcps_len_AtariGo}), and the average depth in the search tree at which new leaves are added (Tables~\ref{tab:grave_dep_AtariGo} and~\ref{tab:mcps_dep_AtariGo}). The three quantities are reported separately for GRAVE and MCPS so that the relative behavior of the two algorithms is directly visible cell by cell. Two patterns are worth noting at a glance: MCPS and GRAVE run at nearly identical playouts per second across all board sizes, and the relative depth gain of MCPS is roughly uniform across sizes, with new leaves added $10$--$25\%$ deeper than under GRAVE at the $2$s budget.

\begin{table}[ht]
\centering
\begin{tabular}{llrrrr}
\hline
Game & Size & 0.25s & 0.5s & 1s & 2s \\
\hline
AtariGo & $5\times 5$ & 218724 & 216393 & 216843 & 226065 \\
AtariGo & $7\times 7$ & 83764 & 80434 & 78242 & 77326 \\
AtariGo & $9\times 9$ & 31193 & 32635 & 32352 & 31105 \\
AtariGo & $11\times 11$ & 14740 & 14956 & 16465 & 17066 \\
AtariGo & $13\times 13$ & 8316 & 8130 & 8436 & 8805 \\
AtariGo & $15\times 15$ & 4874 & 5123 & 4960 & 4915 \\
\hline
\end{tabular}
\caption{GRAVE playouts per second on AtariGo. Time budget per move. Cell = playouts/s per thread.}
\label{tab:grave_pps_AtariGo}
\end{table}

\begin{table}[ht]
\centering
\begin{tabular}{llrrrr}
\hline
Game & Size & 0.25s & 0.5s & 1s & 2s \\
\hline
AtariGo & $5\times 5$ & 227618 & 231783 & 231149 & 223520 \\
AtariGo & $7\times 7$ & 85393 & 84624 & 83082 & 79784 \\
AtariGo & $9\times 9$ & 32977 & 33682 & 34814 & 34262 \\
AtariGo & $11\times 11$ & 15340 & 16283 & 17734 & 18373 \\
AtariGo & $13\times 13$ & 8274 & 8520 & 8656 & 9898 \\
AtariGo & $15\times 15$ & 5106 & 5145 & 5175 & 5244 \\
\hline
\end{tabular}
\caption{MCPS playouts per second on AtariGo. Time budget per move. Cell = playouts/s per thread.}
\label{tab:mcps_pps_AtariGo}
\end{table}

\begin{table}[ht]
\centering
\begin{tabular}{llrrrr}
\hline
Game & Size & 0.25s & 0.5s & 1s & 2s \\
\hline
AtariGo & $5\times 5$ & 4.9 & 4.9 & 4.8 & 4.6 \\
AtariGo & $7\times 7$ & 7.1 & 7.4 & 7.6 & 7.6 \\
AtariGo & $9\times 9$ & 10.7 & 10.2 & 10.4 & 10.8 \\
AtariGo & $11\times 11$ & 15.7 & 15.4 & 14.1 & 13.7 \\
AtariGo & $13\times 13$ & 19.8 & 20.0 & 19.2 & 18.4 \\
AtariGo & $15\times 15$ & 25.4 & 23.9 & 24.6 & 24.8 \\
\hline
\end{tabular}
\caption{GRAVE average playout length on AtariGo. Time budget per move. Cell = avg moves per playout.}
\label{tab:grave_len_AtariGo}
\end{table}

\begin{table}[ht]
\centering
\begin{tabular}{llrrrr}
\hline
Game & Size & 0.25s & 0.5s & 1s & 2s \\
\hline
AtariGo & $5\times 5$ & 4.2 & 4.2 & 4.1 & 4.1 \\
AtariGo & $7\times 7$ & 6.4 & 6.4 & 6.5 & 6.6 \\
AtariGo & $9\times 9$ & 9.6 & 9.4 & 9.1 & 9.2 \\
AtariGo & $11\times 11$ & 14.6 & 13.7 & 12.6 & 12.1 \\
AtariGo & $13\times 13$ & 19.5 & 18.7 & 18.3 & 15.9 \\
AtariGo & $15\times 15$ & 23.9 & 23.5 & 23.1 & 22.8 \\
\hline
\end{tabular}
\caption{MCPS average playout length on AtariGo. Time budget per move. Cell = avg moves per playout.}
\label{tab:mcps_len_AtariGo}
\end{table}

\begin{table}[ht]
\centering
\begin{tabular}{llrrrr}
\hline
Game & Size & 0.25s & 0.5s & 1s & 2s \\
\hline
AtariGo & $5\times 5$ & 7.4 & 7.9 & 8.4 & 9.0 \\
AtariGo & $7\times 7$ & 5.9 & 6.5 & 7.0 & 7.6 \\
AtariGo & $9\times 9$ & 4.4 & 4.9 & 5.5 & 6.1 \\
AtariGo & $11\times 11$ & 3.4 & 3.8 & 4.3 & 4.8 \\
AtariGo & $13\times 13$ & 2.9 & 3.2 & 3.5 & 3.9 \\
AtariGo & $15\times 15$ & 2.4 & 2.8 & 3.1 & 3.4 \\
\hline
\end{tabular}
\caption{GRAVE average new-leaf depth on AtariGo. Cell = avg tree depth at which a new node is added.}
\label{tab:grave_dep_AtariGo}
\end{table}

\begin{table}[ht]
\centering
\begin{tabular}{llrrrr}
\hline
Game & Size & 0.25s & 0.5s & 1s & 2s \\
\hline
AtariGo & $5\times 5$ & 8.4 & 9.0 & 9.6 & 10.0 \\
AtariGo & $7\times 7$ & 6.9 & 7.7 & 8.4 & 9.1 \\
AtariGo & $9\times 9$ & 5.0 & 5.8 & 6.7 & 7.5 \\
AtariGo & $11\times 11$ & 3.7 & 4.3 & 5.2 & 6.0 \\
AtariGo & $13\times 13$ & 3.0 & 3.5 & 4.0 & 4.6 \\
AtariGo & $15\times 15$ & 2.5 & 2.9 & 3.4 & 3.9 \\
\hline
\end{tabular}
\caption{MCPS average new-leaf depth on AtariGo. Cell = avg tree depth at which a new node is added.}
\label{tab:mcps_dep_AtariGo}
\end{table}

\subsection{Hex}

The following tables give the per-cell measurements underlying the Hex experiments. For each board size and time budget, we report the number of playouts completed per second (Tables~\ref{tab:grave_pps_hex} and~\ref{tab:mcps_pps_hex}), the average number of moves in a playout from the new leaf to a terminal state (Tables~\ref{tab:grave_len_hex} and~\ref{tab:mcps_len_hex}), and the average depth in the search tree at which new leaves are added (Tables~\ref{tab:grave_dep_hex} and~\ref{tab:mcps_dep_hex}). The three quantities are reported separately for GRAVE and MCPS so that the relative behavior of the two algorithms is directly visible cell by cell. Two patterns are worth noting at a glance: MCPS overtakes GRAVE in raw playouts per second around the $9\times 9$ board, and MCPS playouts become substantially shorter than GRAVE's as the board grows, falling to around two-thirds of GRAVE's length at $15\times 15$.

\begin{table}[ht]
\centering
\begin{tabular}{llrrrr}
\hline
Game & Size & 0.25s & 0.5s & 1s & 2s \\
\hline
Hex & $5\times 5$ & 446446 & 447809 & 438283 & 425890 \\
Hex & $7\times 7$ & 175196 & 176288 & 168189 & 164351 \\
Hex & $9\times 9$ & 84086 & 82740 & 82297 & 80768 \\
Hex & $11\times 11$ & 35687 & 39805 & 41496 & 40715 \\
Hex & $13\times 13$ & 20380 & 19782 & 23157 & 24233 \\
Hex & $15\times 15$ & 12883 & 12606 & 12659 & 14295 \\
\hline
\end{tabular}
\caption{GRAVE playouts per second on Hex. Time budget per move. Cell = playouts/s per thread.}
\label{tab:grave_pps_hex}
\end{table}

\begin{table}[ht]
\centering
\begin{tabular}{llrrrr}
\hline
Game & Size & 0.25s & 0.5s & 1s & 2s \\
\hline
Hex & $5\times 5$ & 373260 & 381576 & 381487 & 378534 \\
Hex & $7\times 7$ & 154108 & 149815 & 149945 & 145970 \\
Hex & $9\times 9$ & 82690 & 83425 & 78084 & 72962 \\
Hex & $11\times 11$ & 43264 & 43904 & 44910 & 44049 \\
Hex & $13\times 13$ & 23820 & 26331 & 27924 & 28275 \\
Hex & $15\times 15$ & 13620 & 15388 & 17162 & 17688 \\
\hline
\end{tabular}
\caption{MCPS playouts per second on Hex. Time budget per move. Cell = playouts/s per thread.}
\label{tab:mcps_pps_hex}
\end{table}

\begin{table}[ht]
\centering
\begin{tabular}{llrrrr}
\hline
Game & Size & 0.25s & 0.5s & 1s & 2s \\
\hline
Hex & $5\times 5$ & 4.4 & 4.3 & 4.3 & 4.2 \\
Hex & $7\times 7$ & 8.6 & 8.4 & 8.4 & 7.9 \\
Hex & $9\times 9$ & 16.2 & 16.0 & 15.7 & 15.2 \\
Hex & $11\times 11$ & 30.2 & 26.7 & 25.2 & 25.1 \\
Hex & $13\times 13$ & 48.9 & 49.6 & 41.2 & 38.6 \\
Hex & $15\times 15$ & 68.4 & 68.7 & 66.1 & 58.5 \\
\hline
\end{tabular}
\caption{GRAVE average playout length on Hex. Time budget per move. Cell = avg moves per playout.}
\label{tab:grave_len_hex}
\end{table}

\begin{table}[ht]
\centering
\begin{tabular}{llrrrr}
\hline
Game & Size & 0.25s & 0.5s & 1s & 2s \\
\hline
Hex & $5\times 5$ & 4.1 & 4.0 & 4.0 & 3.9 \\
Hex & $7\times 7$ & 7.5 & 7.4 & 7.0 & 6.6 \\
Hex & $9\times 9$ & 13.2 & 12.3 & 12.4 & 12.5 \\
Hex & $11\times 11$ & 21.5 & 20.1 & 18.7 & 18.1 \\
Hex & $13\times 13$ & 37.5 & 31.9 & 28.4 & 27.0 \\
Hex & $15\times 15$ & 59.9 & 50.5 & 43.1 & 40.2 \\
\hline
\end{tabular}
\caption{MCPS average playout length on Hex. Time budget per move. Cell = avg moves per playout.}
\label{tab:mcps_len_hex}
\end{table}

\begin{table}[ht]
\centering
\begin{tabular}{llrrrr}
\hline
Game & Size & 0.25s & 0.5s & 1s & 2s \\
\hline
Hex & $5\times 5$ & 7.1 & 7.6 & 8.1 & 8.5 \\
Hex & $7\times 7$ & 6.1 & 6.7 & 7.1 & 7.4 \\
Hex & $9\times 9$ & 5.3 & 5.7 & 6.1 & 6.6 \\
Hex & $11\times 11$ & 4.6 & 5.1 & 5.5 & 5.9 \\
Hex & $13\times 13$ & 4.2 & 4.6 & 5.0 & 5.5 \\
Hex & $15\times 15$ & 3.9 & 4.3 & 4.6 & 5.1 \\
\hline
\end{tabular}
\caption{GRAVE average new-leaf depth on Hex. Cell = avg tree depth at which a new node is added.}
\label{tab:grave_dep_hex}
\end{table}

\begin{table}[ht]
\centering
\begin{tabular}{llrrrr}
\hline
Game & Size & 0.25s & 0.5s & 1s & 2s \\
\hline
Hex & $5\times 5$ & 7.6 & 8.1 & 8.5 & 8.8 \\
Hex & $7\times 7$ & 7.4 & 8.1 & 8.6 & 9.1 \\
Hex & $9\times 9$ & 6.6 & 7.3 & 8.0 & 8.6 \\
Hex & $11\times 11$ & 5.7 & 6.4 & 7.2 & 7.9 \\
Hex & $13\times 13$ & 5.0 & 5.8 & 6.5 & 7.3 \\
Hex & $15\times 15$ & 4.5 & 5.2 & 5.9 & 6.8 \\
\hline
\end{tabular}
\caption{MCPS average new-leaf depth on Hex. Cell = avg tree depth at which a new node is added.}
\label{tab:mcps_dep_hex}
\end{table}

\subsection{Go}

The following tables give the per-cell measurements underlying the Go experiments. For each board size and time budget, we report the number of playouts completed per second (Tables~\ref{tab:grave_pps_go} and~\ref{tab:mcps_pps_go}), the average number of moves in a playout from the new leaf to a terminal state (Tables~\ref{tab:grave_len_go} and~\ref{tab:mcps_len_go}), and the average depth in the search tree at which new leaves are added (Tables~\ref{tab:grave_dep_go} and~\ref{tab:mcps_dep_go}). The three quantities are reported separately for GRAVE and MCPS so that the relative behavior of the two algorithms is directly visible cell by cell. Two patterns are worth noting at a glance: MCPS and GRAVE run at comparable playouts per second across all board sizes, but the new-leaf depth behaves very differently as the time budget grows --- MCPS deepens its tree steadily with more time, while GRAVE's new-leaf depth is non-monotonic and even decreases with longer budgets on several board sizes.

\begin{table}[ht]
\centering
\begin{tabular}{llrrrr}
\hline
Game & Size & 0.25s & 0.5s & 1s & 2s \\
\hline
Go & $5\times 5$ & 72748 & 72639 & 75655 & 74691 \\
Go & $7\times 7$ & 23971 & 23657 & 23870 & 24147 \\
Go & $9\times 9$ & 7853 & 8102 & 8244 & 8138 \\
Go & $11\times 11$ & 2889 & 3033 & 3116 & 3164 \\
Go & $13\times 13$ & 1381 & 1407 & 1392 & 1370 \\
Go & $15\times 15$ & 723 & 696 & 717 & 679 \\
\hline
\end{tabular}
\caption{GRAVE playouts per second on Go. Time budget per move. Cell = playouts/s per thread.}
\label{tab:grave_pps_go}
\end{table}

\begin{table}[ht]
\centering
\begin{tabular}{llrrrr}
\hline
Game & Size & 0.25s & 0.5s & 1s & 2s \\
\hline
Go & $5\times 5$ & 83347 & 81845 & 76523 & 77114 \\
Go & $7\times 7$ & 24065 & 23709 & 22998 & 23130 \\
Go & $9\times 9$ & 8643 & 8380 & 8373 & 8447 \\
Go & $11\times 11$ & 3393 & 3309 & 3318 & 3276 \\
Go & $13\times 13$ & 1481 & 1516 & 1455 & 1503 \\
Go & $15\times 15$ & 768 & 795 & 799 & 779 \\
\hline
\end{tabular}
\caption{MCPS playouts per second on Go. Time budget per move. Cell = playouts/s per thread.}
\label{tab:mcps_pps_go}
\end{table}

\begin{table}[ht]
\centering
\begin{tabular}{llrrrr}
\hline
Game & Size & 0.25s & 0.5s & 1s & 2s \\
\hline
Go & $5\times 5$ & 7.8 & 7.7 & 7.4 & 7.5 \\
Go & $7\times 7$ & 12.4 & 12.7 & 12.5 & 12.2 \\
Go & $9\times 9$ & 21.4 & 20.9 & 20.7 & 21.2 \\
Go & $11\times 11$ & 32.0 & 31.8 & 32.2 & 31.7 \\
Go & $13\times 13$ & 42.7 & 42.6 & 43.7 & 45.9 \\
Go & $15\times 15$ & 56.3 & 59.6 & 59.2 & 62.8 \\
\hline
\end{tabular}
\caption{GRAVE average playout length on Go. Time budget per move. Cell = avg moves per playout.}
\label{tab:grave_len_go}
\end{table}

\begin{table}[ht]
\centering
\begin{tabular}{llrrrr}
\hline
Game & Size & 0.25s & 0.5s & 1s & 2s \\
\hline
Go & $5\times 5$ & 6.8 & 6.8 & 7.0 & 7.0 \\
Go & $7\times 7$ & 12.0 & 12.2 & 12.4 & 12.2 \\
Go & $9\times 9$ & 19.0 & 19.6 & 19.8 & 19.8 \\
Go & $11\times 11$ & 27.0 & 28.5 & 29.4 & 30.0 \\
Go & $13\times 13$ & 39.6 & 39.3 & 41.4 & 41.2 \\
Go & $15\times 15$ & 52.9 & 51.8 & 52.9 & 54.1 \\
\hline
\end{tabular}
\caption{MCPS average playout length on Go. Time budget per move. Cell = avg moves per playout.}
\label{tab:mcps_len_go}
\end{table}

\begin{table}[ht]
\centering
\begin{tabular}{llrrrr}
\hline
Game & Size & 0.25s & 0.5s & 1s & 2s \\
\hline
Go & $5\times 5$ & 6.6 & 6.9 & 7.3 & 7.7 \\
Go & $7\times 7$ & 5.5 & 5.7 & 6.0 & 6.4 \\
Go & $9\times 9$ & 6.9 & 6.1 & 5.9 & 6.0 \\
Go & $11\times 11$ & 10.4 & 8.5 & 7.0 & 6.4 \\
Go & $13\times 13$ & 11.6 & 12.6 & 10.2 & 8.3 \\
Go & $15\times 15$ & 9.8 & 12.7 & 13.7 & 12.4 \\
\hline
\end{tabular}
\caption{GRAVE average new-leaf depth on Go. Cell = avg tree depth at which a new node is added.}
\label{tab:grave_dep_go}
\end{table}

\begin{table}[ht]
\centering
\begin{tabular}{llrrrr}
\hline
Game & Size & 0.25s & 0.5s & 1s & 2s \\
\hline
Go & $5\times 5$ & 8.9 & 9.9 & 11.0 & 12.2 \\
Go & $7\times 7$ & 8.2 & 8.8 & 9.4 & 10.2 \\
Go & $9\times 9$ & 10.8 & 10.8 & 10.7 & 11.4 \\
Go & $11\times 11$ & 12.2 & 12.8 & 13.3 & 13.5 \\
Go & $13\times 13$ & 11.7 & 13.5 & 14.0 & 14.9 \\
Go & $15\times 15$ & 9.8 & 12.6 & 14.5 & 15.8 \\
\hline
\end{tabular}
\caption{MCPS average new-leaf depth on Go. Cell = avg tree depth at which a new node is added.}
\label{tab:mcps_dep_go}
\end{table}

\subsection{NoGo}

The following tables give the per-cell measurements underlying the NoGo experiments. For each board size and time budget, we report the number of playouts completed per second (Tables~\ref{tab:grave_pps_NoGo} and~\ref{tab:mcps_pps_NoGo}), the average number of moves in a playout from the new leaf to a terminal state (Tables~\ref{tab:grave_len_NoGo} and~\ref{tab:mcps_len_NoGo}), and the average depth in the search tree at which new leaves are added (Tables~\ref{tab:grave_dep_NoGo} and~\ref{tab:mcps_dep_NoGo}). The three quantities are reported separately for GRAVE and MCPS so that the relative behavior of the two algorithms is directly visible cell by cell. Two patterns are worth noting at a glance: every individual measurement is more favorable to MCPS than to GRAVE across all sizes and budgets, with no exceptions or sign-flips, and the absolute depth gain of MCPS over GRAVE grows steadily with the time budget at every board size.

\begin{table}[ht]
\centering
\begin{tabular}{llrrrr}
\hline
Game & Size & 0.25s & 0.5s & 1s & 2s \\
\hline
NoGo & $5\times 5$ & 220512 & 221871 & 219920 & 217558 \\
NoGo & $7\times 7$ & 90589 & 89332 & 87854 & 87427 \\
NoGo & $9\times 9$ & 37928 & 37188 & 36936 & 36451 \\
NoGo & $11\times 11$ & 19503 & 19309 & 18556 & 18487 \\
NoGo & $13\times 13$ & 11344 & 11345 & 11314 & 11111 \\
NoGo & $15\times 15$ & 6653 & 6575 & 6587 & 6572 \\
\hline
\end{tabular}
\caption{GRAVE playouts per second on NoGo. Time budget per move. Cell = playouts/s per thread.}
\label{tab:grave_pps_NoGo}
\end{table}

\begin{table}[ht]
\centering
\begin{tabular}{llrrrr}
\hline
Game & Size & 0.25s & 0.5s & 1s & 2s \\
\hline
NoGo & $5\times 5$ & 230489 & 232577 & 232463 & 229608 \\
NoGo & $7\times 7$ & 93288 & 92005 & 92473 & 91364 \\
NoGo & $9\times 9$ & 38971 & 39033 & 38585 & 38510 \\
NoGo & $11\times 11$ & 20177 & 20112 & 18957 & 19684 \\
NoGo & $13\times 13$ & 11612 & 11679 & 11668 & 11565 \\
NoGo & $15\times 15$ & 6764 & 6806 & 6773 & 6813 \\
\hline
\end{tabular}
\caption{MCPS playouts per second on NoGo. Time budget per move. Cell = playouts/s per thread.}
\label{tab:mcps_pps_NoGo}
\end{table}

\begin{table}[ht]
\centering
\begin{tabular}{llrrrr}
\hline
Game & Size & 0.25s & 0.5s & 1s & 2s \\
\hline
NoGo & $5\times 5$ & 5.4 & 5.4 & 5.3 & 5.2 \\
NoGo & $7\times 7$ & 8.7 & 8.7 & 8.7 & 8.6 \\
NoGo & $9\times 9$ & 12.4 & 12.6 & 12.5 & 12.5 \\
NoGo & $11\times 11$ & 16.9 & 17.0 & 17.2 & 17.3 \\
NoGo & $13\times 13$ & 21.7 & 21.7 & 21.7 & 21.9 \\
NoGo & $15\times 15$ & 27.1 & 27.3 & 27.3 & 27.4 \\
\hline
\end{tabular}
\caption{GRAVE average playout length on NoGo. Time budget per move. Cell = avg moves per playout.}
\label{tab:grave_len_NoGo}
\end{table}

\begin{table}[ht]
\centering
\begin{tabular}{llrrrr}
\hline
Game & Size & 0.25s & 0.5s & 1s & 2s \\
\hline
NoGo & $5\times 5$ & 4.9 & 4.9 & 4.8 & 4.7 \\
NoGo & $7\times 7$ & 8.1 & 8.0 & 7.8 & 7.8 \\
NoGo & $9\times 9$ & 11.7 & 11.6 & 11.4 & 11.3 \\
NoGo & $11\times 11$ & 16.1 & 15.9 & 16.3 & 15.7 \\
NoGo & $13\times 13$ & 20.9 & 20.7 & 20.6 & 20.5 \\
NoGo & $15\times 15$ & 26.4 & 26.0 & 26.1 & 25.9 \\
\hline
\end{tabular}
\caption{MCPS average playout length on NoGo. Time budget per move. Cell = avg moves per playout.}
\label{tab:mcps_len_NoGo}
\end{table}

\begin{table}[ht]
\centering
\begin{tabular}{llrrrr}
\hline
Game & Size & 0.25s & 0.5s & 1s & 2s \\
\hline
NoGo & $5\times 5$ & 6.7 & 7.1 & 7.4 & 7.7 \\
NoGo & $7\times 7$ & 6.0 & 6.3 & 6.7 & 6.9 \\
NoGo & $9\times 9$ & 5.5 & 5.8 & 6.1 & 6.4 \\
NoGo & $11\times 11$ & 5.0 & 5.3 & 5.6 & 5.9 \\
NoGo & $13\times 13$ & 4.6 & 4.9 & 5.2 & 5.5 \\
NoGo & $15\times 15$ & 4.2 & 4.5 & 4.9 & 5.1 \\
\hline
\end{tabular}
\caption{GRAVE average new-leaf depth on NoGo. Cell = avg tree depth at which a new node is added.}
\label{tab:grave_dep_NoGo}
\end{table}

\begin{table}[ht]
\centering
\begin{tabular}{llrrrr}
\hline
Game & Size & 0.25s & 0.5s & 1s & 2s \\
\hline
NoGo & $5\times 5$ & 8.3 & 8.9 & 9.3 & 9.6 \\
NoGo & $7\times 7$ & 7.5 & 8.2 & 8.8 & 9.2 \\
NoGo & $9\times 9$ & 6.6 & 7.3 & 8.0 & 8.6 \\
NoGo & $11\times 11$ & 5.9 & 6.6 & 7.2 & 8.0 \\
NoGo & $13\times 13$ & 5.3 & 6.0 & 6.6 & 7.3 \\
NoGo & $15\times 15$ & 4.8 & 5.4 & 6.0 & 6.7 \\
\hline
\end{tabular}
\caption{MCPS average new-leaf depth on NoGo. Cell = avg tree depth at which a new node is added.}
\label{tab:mcps_dep_NoGo}
\end{table}

\subsection{Wargame}

The following tables give the per-cell measurements underlying the Wargame experiments. For each board size and time budget, we report the number of playouts completed per second (Tables~\ref{tab:grave_pps_wargame} and~\ref{tab:mcps_pps_wargame}), the average number of moves in a playout from the new leaf to a terminal state (Tables~\ref{tab:grave_len_wargame} and~\ref{tab:mcps_len_wargame}), and the average depth in the search tree at which new leaves are added (Tables~\ref{tab:grave_dep_wargame} and~\ref{tab:mcps_dep_wargame}). The three quantities are reported separately for GRAVE and MCPS so that the relative behavior of the two algorithms is directly visible cell by cell. Two patterns are worth noting at a glance: MCPS runs $15$--$33\%$ more playouts per second than GRAVE at every board size, and MCPS playouts are consistently shorter than GRAVE's by an increasing margin, dropping from $\sim 90\%$ of GRAVE's length on $5\times 5$ to $\sim 79\%$ on $13\times 13$.

\begin{table}[ht]
\centering
\begin{tabular}{llrrrr}
\hline
Game & Size & 0.25s & 0.5s & 1s & 2s \\
\hline
Wargame & $5\times 5$ & 139746 & 140123 & 137823 & 133624 \\
Wargame & $7\times 7$ & 49174 & 49154 & 47577 & 46425 \\
Wargame & $9\times 9$ & 24122 & 24990 & 26735 & 26729 \\
Wargame & $11\times 11$ & 12705 & 13207 & 14140 & 15091 \\
Wargame & $13\times 13$ & 7487 & 8073 & 8477 & 8724 \\
\hline
\end{tabular}
\caption{GRAVE playouts per second on Wargame. Time budget per move. Cell = playouts/s per thread.}
\label{tab:grave_pps_wargame}
\end{table}

\begin{table}[ht]
\centering
\begin{tabular}{llrrrr}
\hline
Game & Size & 0.25s & 0.5s & 1s & 2s \\
\hline
Wargame & $5\times 5$ & 179863 & 173686 & 171405 & 169868 \\
Wargame & $7\times 7$ & 58948 & 57282 & 54519 & 53256 \\
Wargame & $9\times 9$ & 29336 & 29911 & 30674 & 30878 \\
Wargame & $11\times 11$ & 16070 & 16441 & 17192 & 18051 \\
Wargame & $13\times 13$ & 9944 & 9934 & 10315 & 10578 \\
\hline
\end{tabular}
\caption{MCPS playouts per second on Wargame. Time budget per move. Cell = playouts/s per thread.}
\label{tab:mcps_pps_wargame}
\end{table}

\begin{table}[ht]
\centering
\begin{tabular}{llrrrr}
\hline
Game & Size & 0.25s & 0.5s & 1s & 2s \\
\hline
Wargame & $5\times 5$ & 8.6 & 8.6 & 8.6 & 8.9 \\
Wargame & $7\times 7$ & 18.8 & 18.9 & 19.2 & 19.5 \\
Wargame & $9\times 9$ & 31.0 & 30.1 & 28.6 & 28.4 \\
Wargame & $11\times 11$ & 47.4 & 46.4 & 44.2 & 41.6 \\
Wargame & $13\times 13$ & 64.5 & 63.1 & 61.5 & 59.9 \\
\hline
\end{tabular}
\caption{GRAVE average playout length on Wargame. Time budget per move. Cell = avg moves per playout.}
\label{tab:grave_len_wargame}
\end{table}

\begin{table}[ht]
\centering
\begin{tabular}{llrrrr}
\hline
Game & Size & 0.25s & 0.5s & 1s & 2s \\
\hline
Wargame & $5\times 5$ & 7.7 & 7.8 & 7.8 & 7.9 \\
Wargame & $7\times 7$ & 16.1 & 16.4 & 16.7 & 16.8 \\
Wargame & $9\times 9$ & 25.5 & 24.7 & 24.0 & 23.6 \\
Wargame & $11\times 11$ & 37.5 & 36.7 & 35.1 & 33.1 \\
Wargame & $13\times 13$ & 50.2 & 50.6 & 49.1 & 47.2 \\
\hline
\end{tabular}
\caption{MCPS average playout length on Wargame. Time budget per move. Cell = avg moves per playout.}
\label{tab:mcps_len_wargame}
\end{table}

\begin{table}[ht]
\centering
\begin{tabular}{llrrrr}
\hline
Game & Size & 0.25s & 0.5s & 1s & 2s \\
\hline
Wargame & $5\times 5$ & 11.9 & 13.2 & 14.5 & 15.9 \\
Wargame & $7\times 7$ & 11.8 & 13.0 & 14.2 & 15.5 \\
Wargame & $9\times 9$ & 10.9 & 12.4 & 13.8 & 14.5 \\
Wargame & $11\times 11$ & 8.9 & 10.3 & 11.7 & 13.1 \\
Wargame & $13\times 13$ & 8.2 & 8.9 & 10.2 & 11.5 \\
\hline
\end{tabular}
\caption{GRAVE average new-leaf depth on Wargame. Cell = avg tree depth at which a new node is added.}
\label{tab:grave_dep_wargame}
\end{table}

\begin{table}[ht]
\centering
\begin{tabular}{llrrrr}
\hline
Game & Size & 0.25s & 0.5s & 1s & 2s \\
\hline
Wargame & $5\times 5$ & 13.4 & 15.0 & 16.5 & 18.1 \\
Wargame & $7\times 7$ & 13.3 & 14.7 & 16.1 & 17.7 \\
Wargame & $9\times 9$ & 11.9 & 13.7 & 15.3 & 16.5 \\
Wargame & $11\times 11$ & 9.7 & 11.2 & 12.9 & 14.5 \\
Wargame & $13\times 13$ & 9.4 & 9.6 & 11.0 & 12.6 \\
\hline
\end{tabular}
\caption{MCPS average new-leaf depth on Wargame. Cell = avg tree depth at which a new node is added.}
\label{tab:mcps_dep_wargame}
\end{table}

\end{document}